\documentclass[conference]{IEEEtran}
\IEEEoverridecommandlockouts
\usepackage{cite}
\usepackage{amsmath,amssymb,amsfonts}
\usepackage{algorithmic}
\usepackage{graphicx}
\usepackage{textcomp}
\usepackage{xcolor}
\usepackage{fancyhdr}
\usepackage{xcolor}
\usepackage{tikz}
\usepackage{multirow}
\usepackage{lipsum}
\usepackage{adjustbox}
\usepackage{booktabs}

\usepackage{siunitx}
\usepackage[a4paper, total={184mm,239mm}]{geometry}
\def\BibTeX{{\rm B\kern-.05em{\sc i\kern-.025em b}\kern-.08em
    T\kern-.1667em\lower.7ex\hbox{E}\kern-.125emX}}

\usepackage{acronym}
\acrodef{EMG}[EMG]{electromyography}
\acrodef{sEMG}[sEMG]{Surface electromyography}
\acrodef{HMI}[HMI]{Human-Machine Interface}
\acrodef{DL}[DL]{Deep Learning}
\acrodef{ML}[ML]{Machine Learning}
\acrodef{SoA}[SoA]{State-of-the-Art}
\acrodef{AP}[AP]{Action Potential}
\acrodef{MUAP}[MUAP]{Motor Unit Action Potential}
\acrodef{MUAPT}[MUAPT]{MUAP Train}
\acrodef{PLI}[PLI]{Power Line Intereference}
\acrodef{MAV}[MAV]{Mean Absolute Value}
\acrodef{WL}[WL]{Waveform Length}
\acrodef{RMS}[RMS]{Root Mean Square}
\acrodef{$k$-NN}[$k$-NN]{$k$-Nearest Neighbors} 
\acrodef{SVM}[SVM]{Support Vector Machine} 
\acrodef{RBF}[RBF]{Radial Basis Function}
\acrodef{RF}[RF]{Random Forest}
\acrodef{HDC}[HDC]{Hyper-Dimensional Computing}
\acrodef{MLP}[MLP]{Multi-Layer Perceptron}
\acrodef{CNN}[CNN]{Convolutional Neural Network}
\acrodef{RNN}[RNN]{Recurrent Neural Network}
\acrodef{LSTM}[LSTM]{Long Short-Term Memory}
\acrodef{TCN}[TCN]{Temporal Convolutional Network}
\acrodef{ReLU}[ReLU]{Rectified Linear Unit}
\acrodef{BN}[BN]{Batch-Normalization}
\acrodef{FC}[FC]{Fully Connected}
\acrodef{MCU}[MCU]{microcontroller unit}
\acrodef{SoC}[SoC]{System on Chip}

\acrodef{MAC}[MAC]{multiply-and-accumulate}
\acrodef{MAE}[MAE]{Mean Absolute Error}
\acrodef{DoF}[DoF]{Degree of Freedom}
\acrodef{DoA}[DoA]{Degree of Actuation}
\acrodef{EMA}[EMA]{Exponential Moving Average}
\acrodef{}[]{}
\acrodef{}[]{}

\renewcommand{\baselinestretch}{0.975}

\usepackage{tikz}
\usepackage{textcomp}
\usepackage[doipre={DOI:~}]{uri}
\usepackage{lipsum}
\newcommand\copyrighttext{%
  \footnotesize \textcopyright 2024 IEEE. Personal use of this material is permitted.  Permission from IEEE must be obtained for all other uses, in any current or future media, including reprinting/republishing this material for advertising or promotional purposes, creating new collective works, for resale or redistribution to servers or lists, or reuse of any copyrighted component of this work in other works.
  }
\newcommand{\copyrightnotice}{%
\begin{tikzpicture}[remember picture,overlay]
\node[anchor=south,yshift=10pt] at (current page.south) {\fbox{\parbox{\dimexpr\textwidth-\fboxsep-\fboxrule\relax}{\copyrighttext}}};
\end{tikzpicture}%
}

\definecolor{somegray}{rgb}{0.5, 0.5, 0.5}
\newcommand{\darkgrayed}[1]{\textcolor{somegray}{#1}}
\makeatletter
\newcommand*\titleheader[1]{\gdef\@titleheader{#1}}
\AtBeginDocument{%
  \let\st@red@title\@title
  \def\@title{%
    \vskip-2.0em
    \bgroup\normalfont\large\centering\@titleheader\par\egroup
    \vskip0.0em\st@red@title}
}

\makeatother

\titleheader{\darkgrayed{This paper has been accepted for publication in the DATE 2024 conference\\\copyright 2024 IEEE.}}

\title{HW-SW Optimization of DNNs for Privacy-preserving People Counting on Low-resolution Infrared Arrays\thanks{This work has received funding from the ECSEL Joint Undertaking (JU) under grant agreement No 101007321. The JU receives support from the European Union’s Horizon 2020 research and innovation programme and France, Belgium, Czech Republic, Germany, Italy, Sweden, Switzerland, Turkey.}}

\begin{document}
\bstctlcite{IEEEexample:BSTcontrol}
\author{\IEEEauthorblockN{
Matteo Risso\IEEEauthorrefmark{1}, Chen Xie\IEEEauthorrefmark{1}, Francesco Daghero\IEEEauthorrefmark{1}, Alessio Burrello\IEEEauthorrefmark{1}, Seyedmorteza Mollaei\IEEEauthorrefmark{1},\\Marco Castellano\IEEEauthorrefmark{2}, Enrico Macii\IEEEauthorrefmark{1}, Massimo Poncino\IEEEauthorrefmark{1}, Daniele Jahier Pagliari\IEEEauthorrefmark{1}}
\IEEEauthorblockA{ 
\IEEEauthorrefmark{1} Politecnico di Torino, Turin, 10129, Italy
\IEEEauthorrefmark{2} ST Microelectronics S.r.l., Cornaredo, 20010, Italy}
\IEEEauthorblockA{Emails: name.surname@polito.it, name.surname@st.it}
}

\maketitle
\copyrightnotice

\begin{abstract}
Low-resolution infrared (IR) array sensors enable people counting applications such as monitoring the occupancy of spaces and people flows while preserving privacy and minimizing energy consumption. Deep Neural Networks (DNNs) have been shown to be well-suited to process these sensor data in an accurate and efficient manner.
Nevertheless, the space of DNNs' architectures is huge and its manual exploration is burdensome and often leads to sub-optimal solutions.
To overcome this problem, in this work, we propose a highly automated full-stack optimization flow for DNNs that goes from neural architecture search, mixed-precision quantization, and post-processing, down to the realization of a new smart sensor prototype, including a Microcontroller with a customized instruction set.
Integrating these cross-layer optimizations, we obtain a large set of Pareto-optimal solutions in the 3D-space of energy, memory, and accuracy.
Deploying such solutions on our hardware platform, we improve the state-of-the-art achieving up to 4.2$\times$ model size reduction, 23.8$\times$ code size reduction, and 15.38$\times$ energy reduction at iso-accuracy.
%
\end{abstract}
\begin{IEEEkeywords}
Deep Learning, Neural Architecture Search, TinyML, MCUs, Smart Sensors
\end{IEEEkeywords}

\section{Introduction}
\label{sec:intro}
\looseness=-1
In the age of pervasive computing, precise counting of people in public and private locations is critical in sectors such as smart buildings and cities, to monitor occupancy and people flows~\cite{rabiee2021multi}.

Classic people counting solutions rely on acquiring a video stream and processing it with a Deep Neural Network (DNN)~\cite{basalamah2019scale, nogueira2019retailnet}.
However, preserving the privacy of individuals is of utmost importance for these applications, and traditional video-based counting systems fail to address this problem by gathering high-resolution visual data~\cite{basalamah2019scale, nogueira2019retailnet}, resulting in ethical and legal concerns, especially for public spaces.
To address this issue, many researchers have proposed the use of low-resolution infrared (IR) sensors~\cite{rabiee2021multi, xie2023efficient}, that capture body heat patterns rather than visual details, allowing for an unobtrusive and anonymous people counting that safeguards sensitive information.

Aside from privacy, achieving low energy consumption is another key objective for people counting systems, e.g., in public areas with no access to the electrical grid. To this end, in-sensor computing is advantageous, as it reduces the dependence on energy-hungry data transfers over a wireless network~\cite{chen2019deep}.
Performing DNN inference on a sensor node, however, introduces a new set of challenges. The complex nature of deep learning models requires intensive computations that may strain the limited capabilities of edge hardware in terms of processing and memory.
%
%

In~\cite{xie2023efficient}, an extensive evaluation of different efficient DNNs led to a rich collection of Pareto-optimal solutions for people-counting on low-resolution IR sensors on the LINAIGE~\cite{linaige} dataset. However, a crucial limitation of~\cite{xie2023efficient} lies in its hand-tuned selection of model architecture configurations. Relying entirely on human expertise may lead to sub-optimal solutions, biased by ``rules-of-thumb'' and designer intuitions.
Furthermore, this approach is severely time-consuming, forcing designers to consider only coarse-grain search spaces. The result is a highly inefficient exploration with a poor ratio between Pareto-optimal solutions and the total number of explored architectures (e.g., 0.8\% in~\cite{xie2023efficient}).
\begin{figure}[t]
  \centering
  \includegraphics[width=1\columnwidth]{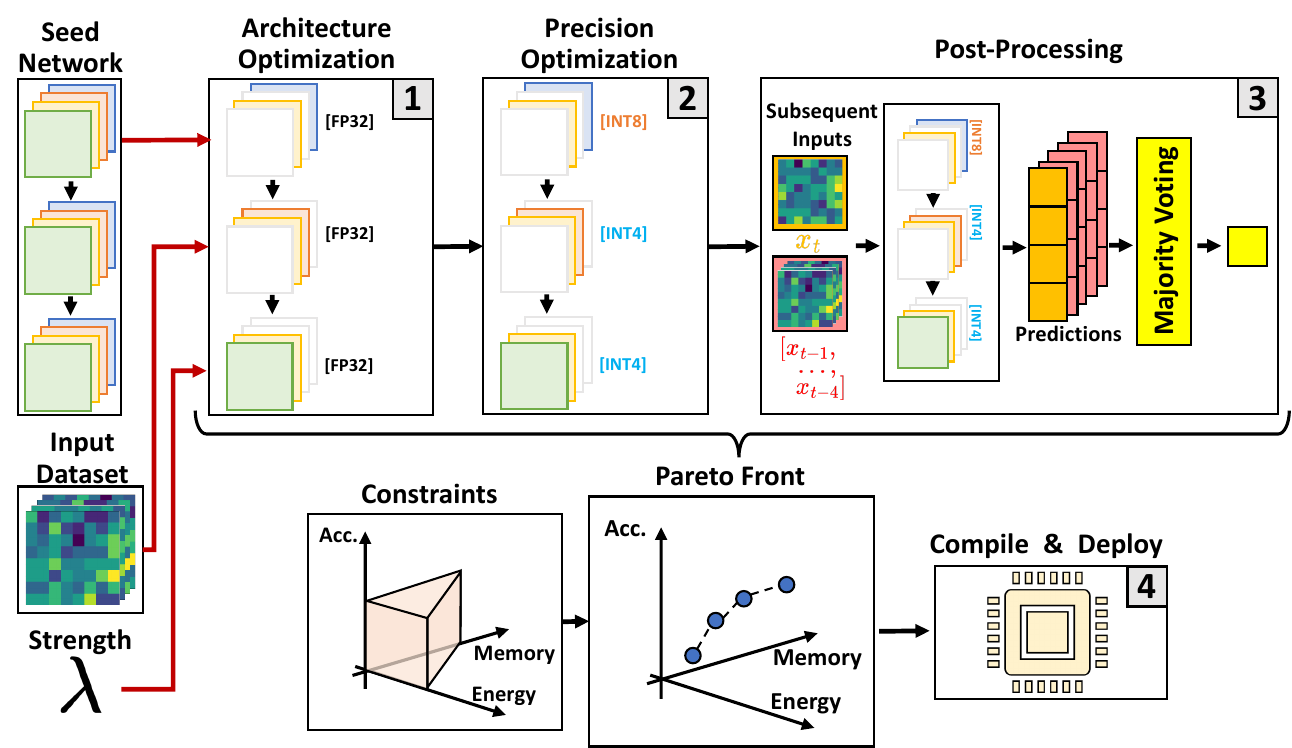}
  \vspace{-0.45cm}
  \caption{Overview of the full-stack optimization flow.}
  \label{fig:flow}
  \vspace{-0.6cm}
\end{figure}

To address these limitations and improve the State-of-the-Art (SotA) for in-sensor IR-based people counting, this work introduces a novel optimization flow that automates the exploration of DNN architectures and quantization formats, along with a novel smart sensor prototype, called MAUPITI, built extending the IBEX~\cite{ibex} RISC-V core to efficiently support low-precision vector operations (Sec.~\ref{sec:hw}).
Fig.~\ref{fig:flow} summarizes the optimization flow. Its main components are:
\begin{itemize}
    \item Complexity-aware Differentiable Neural Architecture Search (DNAS)~\cite{pit} to automatically identify good settings of key DNN hyper-parameters (Sec.~\ref{sec:dnas}).
    \item Mixed-precision quantization of weights and activations of Pareto-optimal DNNs using \texttt{INT4} and \texttt{INT8} data formats (Sec.~\ref{sec:mpq}).
    \item A post-processing technique that improves accuracy exploiting the temporal correlation of subsequent inputs with negligible memory and energy overheads (Sec.~\ref{sec:adaptive}).
    \item Compilation and lightweight inference runtime support for the custom ISA extensions of MAUPITI (Sec.~\ref{sec:deployment}).
\end{itemize}
Thanks to the proposed flow, we were able to obtain a rich collection of Pareto-optimal solutions spanning up to one order of magnitude in memory footprint. 
%
%
%
When compared to SotA, we achieve up to 4.2$\times$ reduction in memory footprint, 23.8$\times$ in code size, and 15.38$\times$ in energy at iso-accuracy.

\section{Background \& Related Work} \label{sec:background_related}
\subsection{Neural Architecture Search (NAS)}
Nowadays, NAS tools represent the go-to solution to help designers in the initial stages of DNN optimization. These tools automatically search for the best DNN architectures among a large number of alternatives described as combinations of different layers and/or hyper-parameters. 
Noteworthy, modern NAS can consider task-specific performance metrics (e.g., accuracy) and non-functional metrics such as memory usage, latency, and energy~\cite{cai2018proxylessnas}.

In particular, Differentiable NAS (DNAS) methods optimize DNNs \textit{while training them}, using gradient descent to solve a relaxed version of the architecture selection problem.
In this way, they significantly reduce the search time with respect to earlier iterative approaches based on Reinforcement Learning or Evolutionary Algorithms (which required 1000s of GPU hours, e.g.~\cite{tan2019mnasnet}), making it comparable to a single training~\cite{liu2018darts, cai2018proxylessnas}.
Some DNAS implementations construct \textit{supernets}, i.e., DNNs with multiple alternative paths, and use the training optimizer to assign a higher probability of being selected to paths that obtain the best accuracy vs cost trade-offs~\cite{liu2018darts,cai2018proxylessnas}.
However, supernets are still very costly to train in terms of GPU memory and latency.
Mask-based DNAS~\cite{pit, wan2020fbnetv2} further reduce optimization costs by exploring a search-space defined by \textit{sub-architectures} contained within a standard DNN, known as the \textit{seed}. Alternative architectures are built by subtraction, removing parts of each layer, such as some channels in a convolution.
In practice, these sub-networks are emulated during training by selectively pruning portions of the seed via \textit{trainable masks}.
While mask-based DNAS strategies are limited to generating networks stemming from the seed, they foster a much more lightweight and fine-grained exploration of the search space~\cite{pit}.
\subsection{Quantization and Mixed-Precision}
Integer quantization is a crucial DNN optimization that replaces floating-point weights and activations with low-bitwidth integers.
Quantization leads to enhancements in model size, speed, and energy efficiency~\cite{jacob_quant}. Furthermore, replacing floating-point computations with integer ones enables the execution of DNNs even on hardware without a Floating Point Unit (FPU).
While quantization can also be applied post-training, simulating its effect with the so-called Quantization-Aware Training (QAT)~\cite{jacob_quant}, is helpful to keep higher accuracy.

Traditional \textit{fixed-precision} quantization employs a uniform bit-width $N_b$ (typically 8 bits) across the model. Recently, however, mixed-precision DNNs using different bit-widths for different portions of the network~\cite{edmips, risso_multiprec} have been shown to provide added benefits in terms of time, memory, and energy, particularly when the underlying hardware supports native sub-byte operations, as in the case of MAUPITI.
\subsection{Privacy-Preserving People Counting}
People counting is important in applications such as smart homes, public security, and pedestrian flow analysis~\cite{rabiee2021multi, giaretta2021people}.
\textit{Instrumented} people counting solutions leverage the Bluetooth or Wi-Fi signals coming from user devices such as wearables and smartphones~\cite{xi2014electronic}. However, their applicability is limited by the strict dependency on individuals' voluntary participation.
\textit{Uninstrumented} solutions, on the other hand, rely on external sensors, such as optical cameras, thermopiles, IR arrays, etc.~\cite{hashimoto1997people}.
Vision-based approaches obtain remarkable results~\cite{basalamah2019scale}, but introduce new important privacy concerns related to storing and processing images that include private user information, such as facial details.
%
%
Low-resolution IR sensors, instead, can detect people without privacy concerns, since they acquire low-resolution thermal images (e.g., 8x8 or 16x16) while allowing much higher counting accuracy with respect to, e.g., single passive IR sensors~\cite{grideye}.

Previous works that exploit IR arrays for people counting can be divided into deterministic (non-data-driven) approaches~\cite{perra2021monitoring, rabiee2021multi, grideye} and Machine Learning (ML)-based methods~\cite{chidurala2021occupancy, bouazizi2022low, en14154542, xie2023efficient}. 
Deterministic algorithms are based on hand-crafted feature-extraction procedures tailored for specific application scenarios.
In general, all these solutions suffer from limited generality to new environments.
For instance, \cite{perra2021monitoring} only counts people entering/exiting a room using doorway-mounted IR arrays. Instead, \cite{grideye, rabiee2021multi} monitor more general environments with ceiling/side wall-mounted IR arrays. However, \cite{rabiee2021multi} relies on a network of multiple IR sensors, while \cite{grideye} is based on a single low-resolution IR array, but suffers from low accuracy as analyzed in~\cite{xie2023efficient}. 

Among ML-based solutions, \cite{chidurala2021occupancy} analyzed multiple classic models such as Random Forests (RFs), Support Vector Machines (SVMs) etc., showing that RFs achieve the highest accuracy on low-resolution (8x8) arrays, but also that results are strongly dependent on the selected hand-crafted feature set.
DNNs are, therefore, a promising way to get rid of manual feature extraction while achieving even higher accuracy.
\cite{bouazizi2022low, en14154542} apply different types of DNNs, including Convolutional Neural Networks (CNNs), Feedforward Neural Networks (FNNs), or Gated Recurrent Units (GRUs) to this task. While proving the superior accuracy of deep learning, these solutions are deployed only on high-performance, energy-hungry, PC-class processors~\cite{chidurala2021occupancy, bouazizi2022low}, and/or utilize relatively high-resolution arrays (e.g., 80x60), which reduces privacy protection.
More recently, \cite{xie2023efficient} provided an extensive comparative analysis of 6 efficient DNN model families for people counting based on a single ultra-low-resolution (8x8) IR array, deploying the results on a commercial Microcontroller (MCU). However, the architectural exploration in \cite{xie2023efficient} was manual and coarse-grained, limiting the possible Pareto-optimal solutions that could be found. Moreover, the MCU considered in \cite{xie2023efficient} did not include specific hardware features to accelerate the execution of low-precision DNNs, which may lead to further efficiency benefits.
\section{Methods}
\label{sec:methods}
This section details our full-stack optimization flow and our novel HW platform, comprising a low-power IR sensor and a RISC-V core optimized for low-precision DNN inference.

The goal of the flow, depicted in Fig.~\ref{fig:flow}, is to obtain a rich set of DNN models for IR-based people counting, offering diverse trade-offs in terms of task performance and HW cost. The latter is measured in terms of n. of parameters (a proxy for memory), or n. of Multiply and Accumulate (MAC) operations (a proxy for energy).
The three inputs to the flow are the training dataset, a seed DNN that acts as ``blueprint" to generate all output solutions, and a scalar parameter $\lambda$ used to control the trade-off between task performance and HW-cost.
Note that the novelty of our method does not reside in the single steps, but in their concatenation into a full-stack flow and application to the IR-based people counting task.
\subsection{Software Optimization Flow}\label{sec:flow}
\subsubsection{Architecture Optimization} \label{sec:dnas}
\begin{figure}[t]
  \centering
  \includegraphics[width=.85\columnwidth]{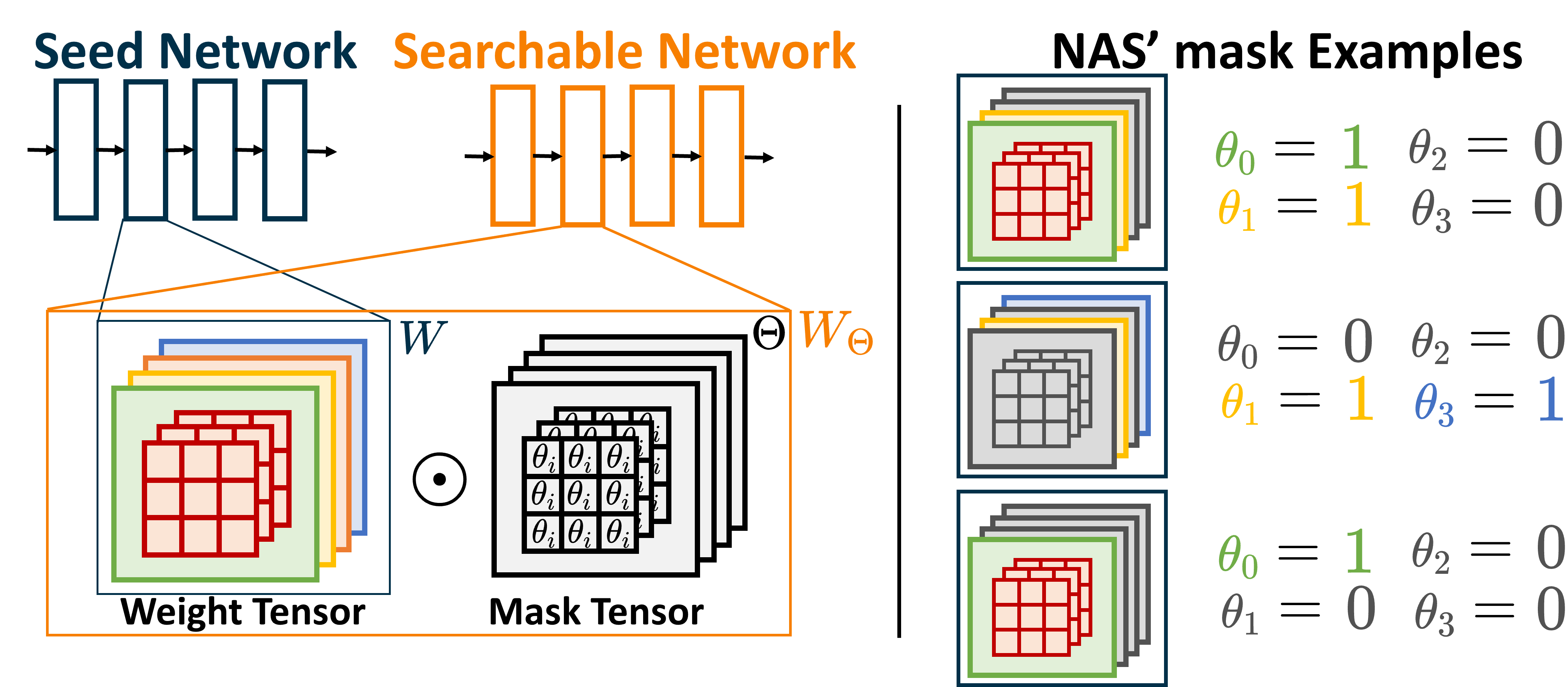}
  \vspace{-0.25cm}
  \caption{Left: The PIT mask-based DNAS scheme. Right: three example results from the search procedure on a layer with four filters.}
  \label{fig:dnas}
  \vspace{-0.5cm}
\end{figure}
The architecture optimization step is based on the PIT~\cite{pit} mask-based DNAS, as implemented in~\cite{pagliari2023plinio}. PIT starts from a seed CNN and explores sub-architectures contained in it by structurally pruning the output channels/features of convolutional and linear layers.
We selected a mask-based approach due to its lower memory and time overhead compared to other NAS tools~\cite{tan2019mnasnet, liu2018darts, cai2018proxylessnas}.
Moreover, such an approach allows to leverage SotA seeds as starting points for the exploration. In particular, we use the DNNs found manually in~\cite{xie2023efficient} as a starting point, demonstrating how fine-grained exploration results in new optimization opportunities.

As schematized in Fig.~\ref{fig:dnas}, PIT considers all convolutional/linear layers of the seed and couples, through Hadamard product $\odot$, each output channel $c$ of the weight tensor $W$ with a binarized trainable mask parameters $\theta_c$:
\begin{equation}
    W^{c}_{\Theta} = W^{c} \odot \mathcal{H}(\theta_c)
\end{equation}
where $\mathcal{H}$ is the Heavised step function that binarizes $\theta$ and $W^{c}$ denotes a specific slice of the weight tensor on the output channel dimension. 

The obtained network is then inserted in a standard training loop where both weights $W$ and masks $\theta$ are trained to minimize the following objective function:
\begin{equation} \label{eq:dnas}
    \min_{W, \theta} \mathcal{L}(W; \theta) + \lambda \mathcal{C}(\theta)
\end{equation}
In (\ref{eq:dnas}) $\mathcal{L}$ is a standard task-specific loss function (e.g., cross-entropy) and $\mathcal{C}$ is a differentiable model of a HW cost metric, e.g., the memory-footprint or the number of MACs. $\lambda$ is a strength parameter that controls the balance between task performance and cost. The higher the value of $\lambda$, the higher the importance given to the minimization of the HW-related cost.
Each value of $\lambda$ will correspond to a specific architecture in the task-performance vs. cost space as detailed in Sec.~\ref{sec:arch_res}.
\subsubsection{Precision Optimization} \label{sec:mpq}
Starting from the optimized architectures obtained with the NAS, we then explore DNN quantization by decreasing the precision of both data and operations from the standard \texttt{FLOAT32} to integer.
We first fold Batch-Normalization (BN) operations with previous convolutional/linear layers, to reduce the total number of operations.
Then, each floating point tensor $T$ (both activations and weights) is quantized to an integer precision $N_b$, by means of an \textit{affine transformation}:
\begin{equation}
    \hat{T} = \text{round}(\frac{T - \alpha}{\beta - \alpha} (2^{N_b} - 1) )
\end{equation}
where $\hat{T}$ is the integer image of $T$ and $[\alpha, \beta]$  can either represent the extremes of the variation range for $T$ or can be learned during training~\cite{choi2018pact}.
In this work, we consider a range-based quantization for weights and a learnable one for activations.
Moreover, we apply QAT, which is particularly important to recover part of the floating-point accuracy when BN folding and sub-byte quantization are employed.

Our preliminary QAT experiments showed a too high accuracy drop for precisions lower than \texttt{INT4} precision. This led to the design choice of supporting only \texttt{INT4} and \texttt{INT8} precisions in the MAUPITI hardware (see Sec.~\ref{sec:hw}).
We use a mixed-precision scheme~\cite{edmips} with a layer-wise granularity to explore different precision assignments, targeting the aforementioned formats. Our method differs with respect to standard mixed-precision, which independently assigns a bit-width to weights and activations. In fact, in order to minimize HW overheads, MAUPITI only supports 4x4-bit and 8x8-bit vectorial MAC operations, i.e., the precision assignment can be different for different layers, but must be the same for weights and activations of the same layer.
Given the limited search space (detailed in Sec.~\ref{sec:results}), we run a complete exploration of all the possible alternatives using the QAT capabilities of~\cite{pagliari2023plinio}.
\subsubsection{Post-Processing} \label{sec:adaptive}
The third step of the flow of Fig.~\ref{fig:flow} consists of a simple yet effective post-processing technique based on majority voting, i.e., \textit{mode inference}.
The rationale is to take advantage of multiple classification results to generate final predictions with lower variance, by exploiting the temporal correlations among subsequent frames.
In practice, the same DNN classifier is applied independently to each input. Then, the final output is built as the most frequently predicted class over a sliding window of recent frames, thus filtering out sporadic mispredictions.
This solution is particularly interesting from the edge computing perspective, having approximately the same memory cost as a single-frame classifier.
A similar approach was considered in~\cite{xie2023efficient}, but in that paper, the network was executed multiple times, discarding previous predictions, thus incurring a large latency and energy cost.
In our post-processing, we avoid any re-computation by simply storing previous predictions in a FIFO data structure, thus bringing latency and energy overheads close to 0 too.

\subsection{The MAUPITI Platform}
\subsubsection{Complete System} \label{sec:full_hw}
\begin{figure}[t]
  \centering
    \includegraphics[width=.7\columnwidth]{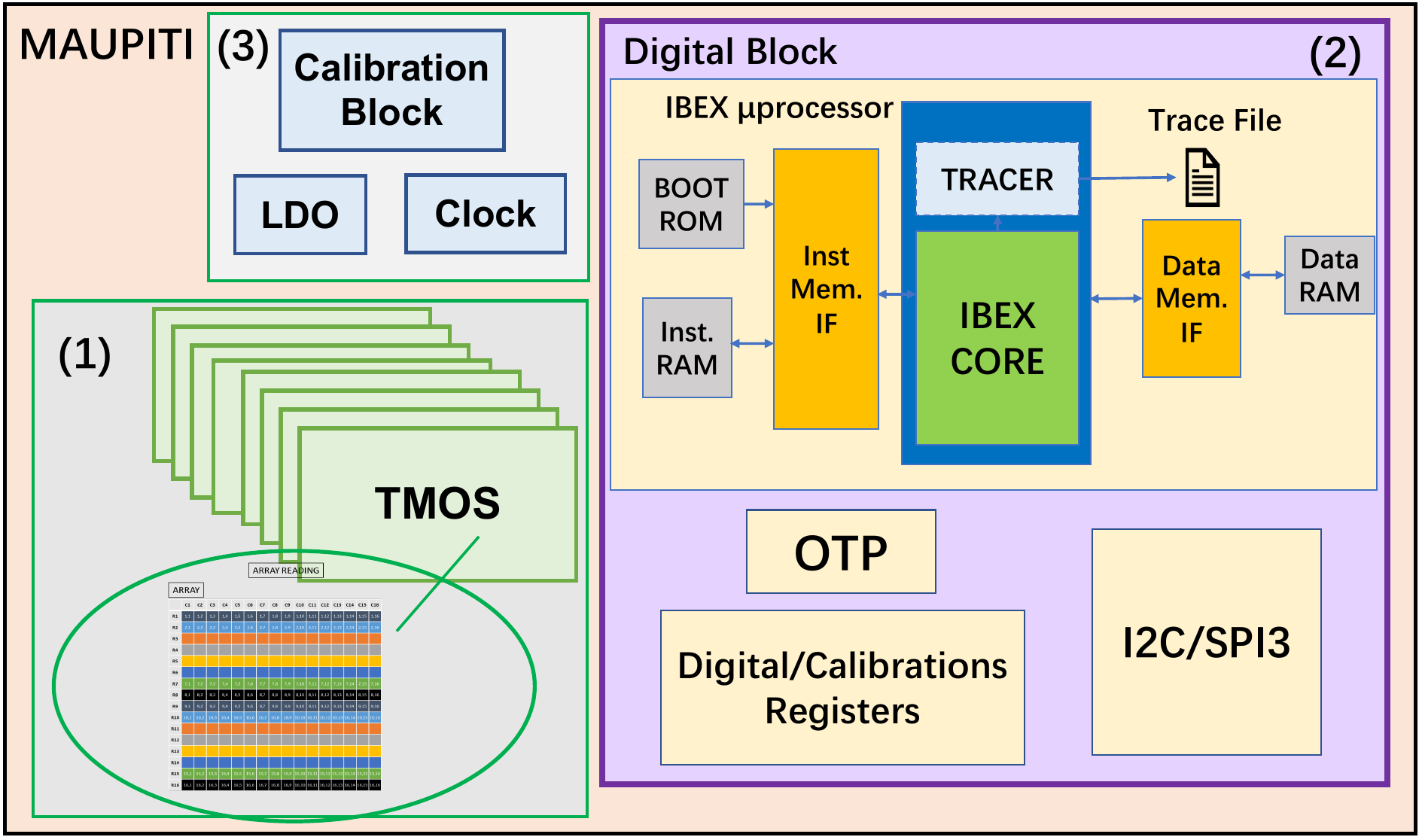}
  \vspace{-0.25cm}
  \caption{The complete MAUPITI System.}
  \label{fig:complete_system}
\end{figure}
\begin{figure}[t]
  \centering
  \includegraphics[width=0.7\columnwidth]{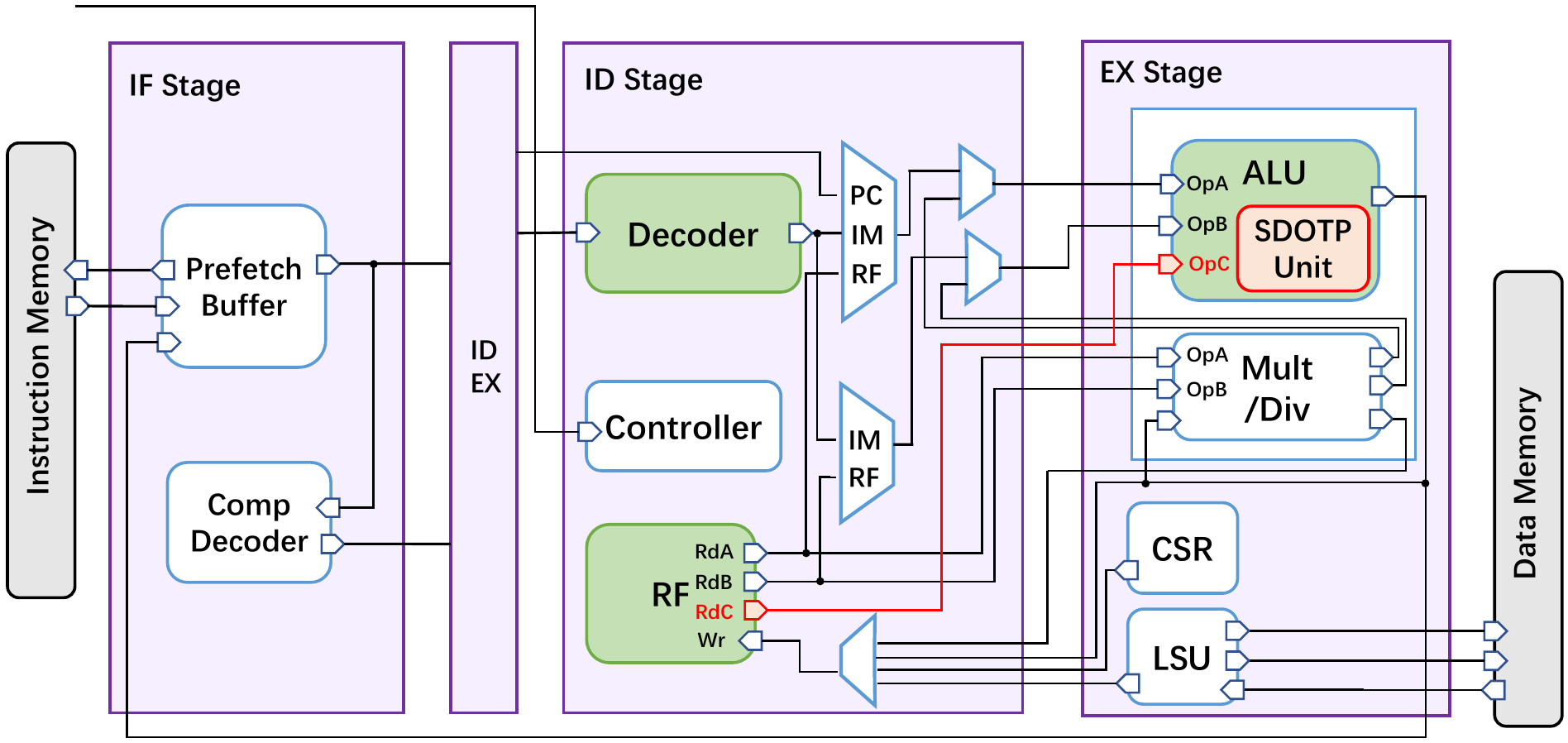}
  \vspace{-0.25cm}
  \caption{Customized IBEX RISC-V core.}
  \label{fig:hw}
  \vspace{-0.4cm}
\end{figure}
%

Fig.~\ref{fig:complete_system} depicts the complete MAUPITI system, which has been taped out in 130nm CMOS technology, is clocked at 20MHz, and can capture thermal images with a frame rate of 10 Frames Per Second (FPS).
The chip includes three main components. A 16x16 array of thermal MOSFET (TMOS) sensors (1), sensitive to the heating effects of infrared radiation, with 8 parallel analog front-end processing chains, each able to acquire one row of pixels, thus enabling a two-step acquisition of a complete frame. Each TMOS draws $\approx1\mu A@2.4V$, resulting in a total consumption of 0.62mW for the array. A digital processing block (2) which contains a customized IBEX core (highlighted in green) with its instruction and data memory (16KB each), and the corresponding interfaces, plus a boot ROM. Moreover, the block also includes an 80B One-Time Programmable Memory (OTP), an instruction tracer, calibrations registers for the TMOS sensors, and I2C and SPI3 communication Interfaces. Overall, the digital block consumes $\approx$0.9mW of power in FF conditions.
Lastly, the system includes standard circuitry blocks (3) for clock and reset management, voltage regulation, etc. 

\subsubsection{Core Customization}\label{sec:hw}
The IBEX core has been customized to support quantized DNNs using \texttt{INT4} and \texttt{INT8} data formats, with the addition of an efficient arithmetic unit supporting low bit-width integer Single-Instruction-Multiple-Data (SIMD) Sum of Dot Product (SDOTP) operations.
Figure \ref{fig:hw} depicts the integration of the SDOTP unit into the IBEX pipeline. Hardware blocks highlighted in green have been modified, while the orange one is entirely new. In detail, the SDOTP unit performs a MAC operation between two 32-bit registers (RS1 and RS2), interpreting their content either as four 8-bit or as eight 4-bit \textit{signed} values, and an additional 32-bit register (RD) is used as input and output of the accumulation. To achieve single-cycle latency, we implement the dot product using four independent 8-bit multipliers, and eight 4-bit multipliers, followed by an adder tree to sum up the partial products and the additional 32-bit operand. Replicating the multipliers for the two bitwidths, instead of sharing them, moves the new block out of the core's critical path, at the cost of an acceptable area increase~\cite{garofalo2021xpulpnn}.
The other key hardware modifications are in the Decoder, extended to support the new opcodes, and in the Register File (RF). The SDOTP uses RD both as source and destination, while the vanilla IBEX ALU only supports two input operands. Thus, we add a third input OpC to the ALU, which always takes data directly from the register file, and the MAC result is then written onto the same register. Accordingly, we must add a read port RdC to the RF, as shown in Figure~\ref{fig:hw}.

With respect to~\cite{garofalo2021xpulpnn}, a SotA mixed-precision ISA extension for RISC-V MCUs, our solution focuses much more on containing area (i.e., cost) overheads, sacrificing some performance and flexibility, since we target a low-cost smart sensors application. To this end, we do not implement unsigned variants of the SDOTP, nor 8x4-bit or 4x8-bit versions, to support different precisions for weights and activations. Similarly, we do not implement 2-bit SDOTP since our QAT experiments showed that 2-bit precision causes strong accuracy degradations. We only support register addressing mode for the source operands (thus requiring separate loads before a dot product), and we do not implement combined MAC\&Load. Lastly, we also avoid having separate instructions for simple dot product (DOTP), since those can be reduced to SDOTP by setting the third input register RD to 0. All these simplifications significantly reduce our area overhead to less than 7\% w.r.t. the vanilla IBEX core. In contrast, \cite{garofalo2021xpulpnn} reported a 17.5\% core area increase despite starting from a more complex baseline core.

\subsubsection{Deployment Toolchain} \label{sec:deployment}
To fully exploit the customized core, we added support for the new SDOTP instructions to the Gnu Compiler Collection (GCC) toolchain for RISC-V targets with the instruction set riscv32-imc.
We then developed a minimal set of optimized kernels to implement the required DNN layers, largely inspired from~\cite{pulp-nn-mixed}, the SotA library for mixed-precision deep learning on RISC-V MCUs. In particular, we developed convolutional kernels taking as input weights/activations at 4/8 bit and writing their re-quantized output again on 4/8 bit. This enables a seamless integration with the quantized architectures found with the flow of Sec.~\ref{sec:flow}.
Aside from convolution, we also have kernels for 2D max-pooling, whereas we avoid a dedicated implementation of linear (fully-connected) layers by re-using the convolutional kernels in the corner case of a 1x1 filter and 1x1 input/output feature maps, thus minimizing the code size of our runtime.

\section{Experimental Results}
\label{sec:results}
\subsection{Setup}
We test our flow on the open-source LINAIGE~\cite{linaige} dataset, which specifically targets the people counting task using $8 \times 8$ IR array sensor data. We use this dataset as it is the largest publicly available one, although the resolution is lower than that of the MAUPITI sensor. Collecting our own dataset will be part of our future work.
The dataset contains 25110 labeled samples, split into 5 sessions (collected in different environments). Each sample is labeled with the number of people in the field of view, ranging from 0 to 3.
In all experiments, the dataset is employed with the same training hyper-parameters and by following the \textit{leave-one-session-out} cross-validation (CV) scheme described in~\cite{xie2023efficient} with Session 1 (the largest) always kept in the training set and Sessions 2, 3, 4, 5 rotated as test-sets.
Namely, we train for 500 epochs using the Adam Optimizer to optimize a cross-entropy loss, with a learning rate of 0.001 and a batch size of 128.
We apply NAS and QAT only on Session 1 data, then apply QAT and fine-tune the found models on the training fold (which includes all sessions but one), and test on the left-out session.
Models' performance is evaluated through the average of recall or Balanced Accuracy Score (BAS).
Differently from~\cite{xie2023efficient}, which runs the experiment only one time, we repeated them ten times with different random seeds. Each result is reported with mean and variance, thus providing a better statistical characterization.

As seed for the DNAS, we use the largest CNN configuration considered in~\cite{xie2023efficient}, which includes a feature-extractor composed of two convolutional layers both with a $3 \times 3$ kernel, a stride of 1, and 64 output channels with a max-pooling layer in between. The network is concluded by two linear layers respectively with 64 and 4 output features. The two convolutions are followed by BN, and all layers except the output one use ReLU as non-linearity.
Taking the largest configuration of~\cite{xie2023efficient}, we aim to demonstrate that with our flow, we are able to match or improve the SotA thanks to a finer-grain, automated search.

All the code is written using Python 3.9, PyTorch 1.13.1 and employs the PLiNIO~\cite{pagliari2023plinio} library to implement NAS and QAT. 
\subsection{Architecture and Precision Space Exploration} \label{sec:arch_res}
\begin{figure}[t]
  \centering
  \includegraphics[width=0.6\columnwidth]{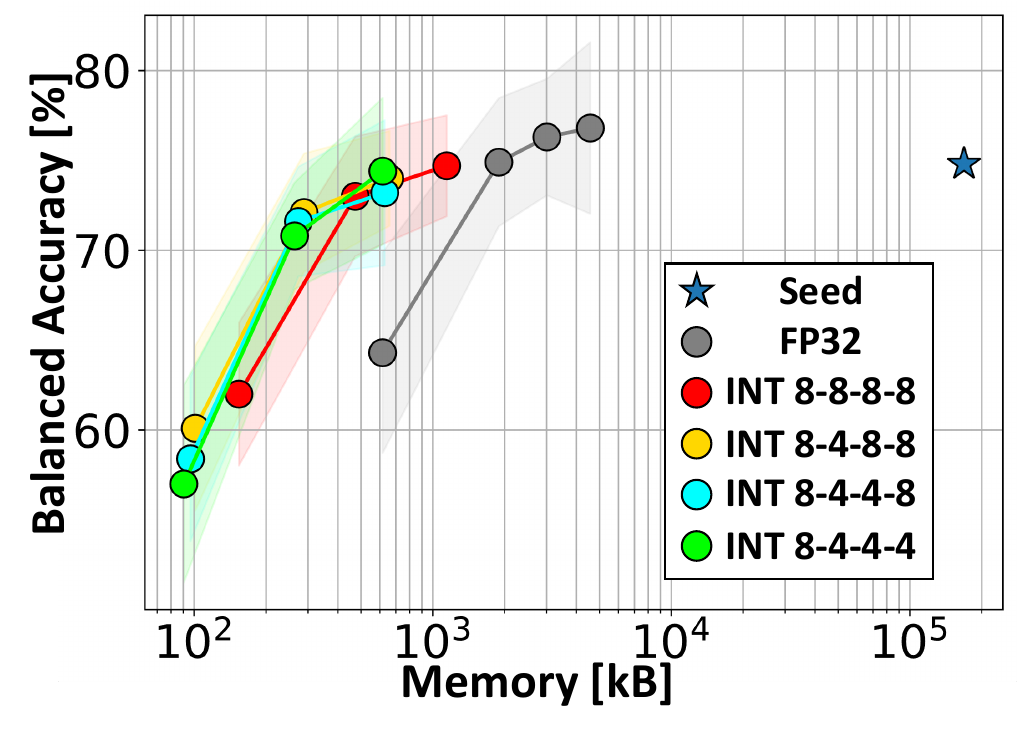}
  \vspace{-0.45cm}
  \caption{Architecture and Precision Search Space exploration results. Different colors encode the different precisions' configurations.}
  \label{fig:arch_prec_expl}
  \vspace{-0.4cm}
\end{figure}
The input of our flow is highlighted with a blue star in Fig.~\ref{fig:arch_prec_expl} in a BAS vs memory plane.
Starting from such point and by applying the PIT~\cite{pit} DNAS with different values of strength $\lambda$ and using the number of parameters as cost $\mathcal{C}$ (See Eq.~\ref{eq:dnas}), we obtain the grey Pareto front. Noteworthy, when compared to the seed, we achieve up to 89$\times$ memory and 26.7$\times$ N. of MACs reduction and iso-BAS.

The next step of the flow is the mixed-precision quantization of the \texttt{FLOAT32} results obtained with PIT. 
The results are summarized in Fig.~\ref{fig:arch_prec_expl} with colored circles, where each color encodes a specific precision combination for the four considered layers.
We only report solutions with the first layer quantized at 8-bit because using \texttt{INT4} to quantize the inputs led to severe accuracy degradations and resulted in sub-optimal solutions.
Moreover, the ``INT 8-8-4-8" is not reported because it is never part of the overall Pareto frontier.
Thanks to quantization, we improve the \texttt{FLOAT32} front up to 2.3$\times$ in terms of memory while also improving the BAS up to 6.5\% w.r.t. the left-most point of the grey curve.
Overall, the memory and MACs reduction w.r.t. the seed reach 147$\times$ and 234$\times$ at iso-BAS.
\subsection{Post-Processing Results}
\begin{figure}[t]
  \centering
  \includegraphics[width=\columnwidth]{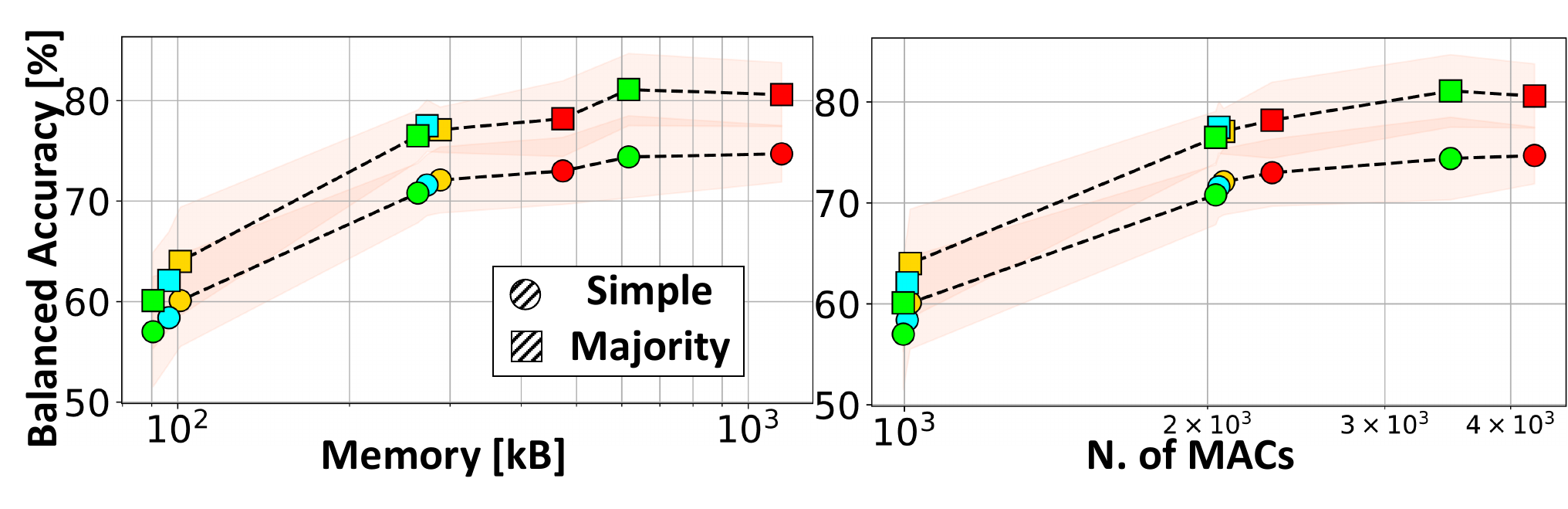}
  \vspace{-0.8cm}
  \caption{Comparison of Pareto frontiers with and without post-processing.}
  \label{fig:post_proc_res}
  \vspace{-0.4cm}
\end{figure}
Fig.~\ref{fig:post_proc_res} shows the results of applying the proposed post-processing scheme on the best networks obtained with architecture and precision explorations. The circles correspond to the global Pareto fronts obtained from the NAS and QAT phases. For instance, the circles' curve in the leftmost plot is obtained merging the optimal points from all quantized curves in Fig.~\ref{fig:arch_prec_expl}. The squares are the results of applying post-processing to the outputs of those DNNs.
The left and right plots show how the majority-voting technique represents a plug-and-play strategy to improve the Pareto-frontier in the BAS vs. Memory and BAS vs. N. of MACs spaces. We considered a sliding window composed of the predictions on 5 subsequent frames, which demonstrated to be the most effective one on the dataset.
Majority voting introduces a negligible delay in detecting people count changes equal to half of the window length (assuming all-correct predictions).
In exchange, it achieves up to 6.7\% BAS improvement at iso-memory and iso-MACs.
\subsection{State-of-the-Art Comparison}
\begin{figure}[t]
  \centering
  \includegraphics[width=\columnwidth]{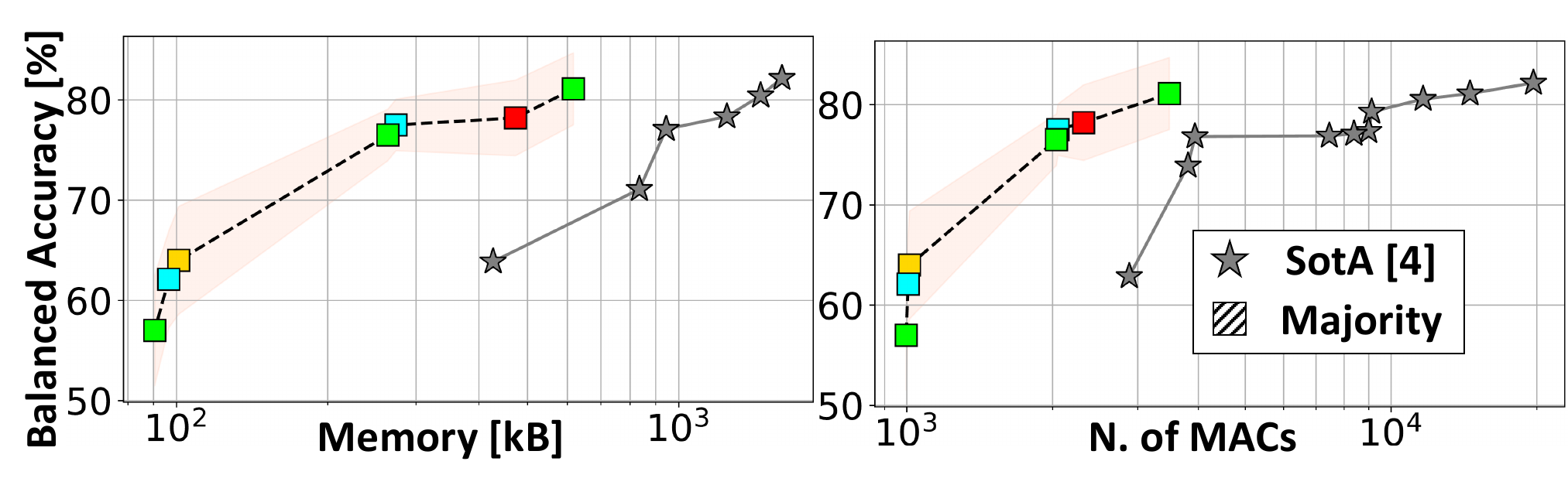}
  \vspace{-0.8cm}
  \caption{State-of-the-Art Comparison.}
  \label{fig:sota-comp}
  \vspace{-0.6cm}
\end{figure}
Fig.~\ref{fig:sota-comp} compares the results obtained with our proposed pipeline, and the SotA results presented in~\cite{xie2023efficient}.
On the left-most plot, we compare the results in the BAS vs. Memory space, and on the right, in the BAS vs N. of MACs space.
As shown, \cite{xie2023efficient} achieved a slightly higher maximum BAS (+0.9\%), which, however, is well within the standard deviation range of our most accurate model.
On the other hand, when targeting a BAS higher than 80\%, we obtain models up to $2.4 \times$ smaller and requiring 3.3$\times$ fewer MACs with respect to the SotA. Similarly, when comparing against the most memory-efficient DNN of \cite{xie2023efficient}, and the one with the lowest number of MACs (the left extremes of the two grey curves), our flow produces models that are respectively 4.2$\times$ smaller and require 2.9$\times$ fewer MACs for the same BAS.

\subsection{Embedded Deployment Results}
\begin{table}[ht]
\footnotesize
\caption{Deployment results.}\label{tab:deployment}
\vspace{-0.2cm}
\centering
\begin{adjustbox}{width=.8\columnwidth}
\begin{tabular}{clccc}
\multicolumn{1}{l}{\textbf{Model}} &
  \textbf{Platform} &
  \multicolumn{1}{l}{\textbf{Code {[}B{]}}} &
  \multicolumn{1}{l}{\textbf{Data {[}B{]}}} &
  \textbf{Energy {[}$\mu$J{]}} \\ \toprule
\multirow{3}{*}{\textbf{Top}}   & STM32 & 22840 & 10420 & 9.381\\
                                & IBEX & 3476 & 1104 & 6.003\\
                                & \textbf{MAUPITI} & 4152 & 1104 & 4.927\\ \midrule
\multirow{3}{*}{\textbf{- 5\%}} & STM32 & 22970 & 9060 & 6.854\\
                                & IBEX & 3384 & 648 & 5.005\\
                                & \textbf{MAUPITI} & 4052 & 648 & 4.525\\ \midrule
\multirow{3}{*}{\textbf{Mini}}  & STM32 & 22950 & 8410 & 5.640\\
                                & IBEX & 2700 & 416  & 4.342\\
                                & \textbf{MAUPITI}  & 3208 & 416 & 4.067\\ \bottomrule
\end{tabular}
\vspace{-0.8cm}
\end{adjustbox}
\end{table}
Table~\ref{tab:deployment} reports the deployment results of the top scoring model (\textit{Top}), the smallest model in terms of memory with a maximum drop in accuracy smaller than 5\% (\textit{-5\%}), and the smallest overall (\textit{Mini}). Noteworthy, in our case, the latter corresponds also to the models with the lowest MACs.
For the aforementioned architectures, we report the code size (Code), the memory occupation (Data), and the energy consumption per inference on three different platforms. 
Namely, we compare the proposed MAUPITI platform with an unmodified IBEX core, using no custom instructions. For these two targets, we use our deployment toolchain with identical compilation flags.
Moreover, we compare MAUPITI to an off-the-shelf MCU solution, i.e., an STM32L4R5 core with models deployed on 8bits only using the proprietary X-CUBE-AI toolchain, which does not support mixed-precision.
%
When comparing with IBEX, the \textit{Top} model, achieves the largest gains in terms of energy.
MAUPITI, at the cost of 7\% area overhead, and despite a 2.2\% post-synthesis power overhead, achieves up to 17.9\% energy reduction.
Note that the efficiency benefits of the low-precision SIMD in MAUPITI are limited by the small geometry of the considered layers, namely the small number of output channels, and would be higher for larger DNNs.
%
%
%
%

Concerning memory, we see a code size increase compared to the vanilla IBEX core. We have the largest increase (676 B) with the \textit{Top} model.
The reason is the more complex logic of MAUPITI kernels, with the baseline versions repeating a regular read-and-multiply pattern independently from the layer hyper-parameters. On the other hand, SIMD instructions read and perform operations on chunks of inputs, thus having to manage non-multiple-of-4(8) dimensions as ``leftovers''.

Thanks to our lightweight runtime, when comparing with STM32 solutions, MAUPITI shows up to $6.78 \times$/$20.22 \times$ code-size and data reduction. Even the most accurate network can easily fit the 16 kB of code memory and 16 kB of data memory available on the chip.
STM32 executes DNNs up to $9 \times$ faster than MAUPITI. This is partly due to the higher frequency of the core (120MHz versus the 20 MHz of MAUPITI), partly to the different ISA, and partly to the optimizations included in X-CUBE-AI, such as max-pooling fusion. Nevertheless, such reduction in latency is paid with a huge increase of power consumption of $13.2 \times$, which makes MAUPITI up to 1.4x-1.9x more efficient in terms of energy.

Finally, when comparing with the deployment results of~\cite{xie2023efficient} we note that their smallest solution (leftmost grey star in Fig.~\ref{fig:sota-comp}) was reported to have a code size 23.8$\times$ higher and to consume 15.38$\times$ more energy compared to our equally accurate DNN (third square from the left in Fig.~\ref{fig:sota-comp}).
Conversely, when comparing our \textit{Top} model with the one in~\cite{xie2023efficient} (rightmost grey star in Fig.~\ref{fig:sota-comp}) we achieve 69$\times$ code size reduction and 24.4$\times$ energy reduction with a small BAS drop of 0.9\%.
\section{Conclusions}
\label{sec:conclusions}
The yet accurate yet privacy-preserving monitoring of people flows in smart buildings and cities contexts represents key requirements.
This work effectively proposes a full-stack optimization pipeline, from software down to hardware, that enables such application to use low-resolution IR arrays and perform DNN inference at the edge.
The proposed flow is able to improve the SotA with up to 4.2$\times$ memory reduction, 23.8$\times$ code-size reduction, and 15.38$\times$ energy reduction at iso-accuracy.

\tiny
\bibliographystyle{IEEEtran}
\bibliography{bstctl,references}

\end{document}